%% file: main.tex
\documentclass[10pt,twocolumn,letterpaper]{article}

\usepackage[pagenumbers]{cvpr} 

\input{preamble}

\definecolor{cvprblue}{rgb}{0.21,0.49,0.74}
\usepackage[pagebackref,breaklinks,colorlinks,allcolors=cvprblue]{hyperref}

\def\paperID{} 
\def\confName{CVPR}
\def\confYear{2026}

\title{VGent: Visual Grounding via Modular Design for \\ Disentangling Reasoning and Prediction}

\author{Weitai Kang$^{1}$~~~~Jason Kuen$^{2}$~~~~Mengwei Ren$^{2}$~~~~Zijun Wei$^{2,}$\footnotemark[1]~~~~~Yan Yan$^{1}$~~~~Kangning Liu$^{2,}$\footnotemark[2] \\
$^1$University of Illinois Chicago~~~~$^2$Adobe
}

\begin{document}
\maketitle
\input{sec/0_abstract}    
\renewcommand{\thefootnote}{\fnsymbol{footnote}}
\footnotetext[1]{Direction Lead \quad \textsuperscript{\dag}Project Lead}
\input{sec/1_intro}
\input{sec/2_related}

\input{sec/3_method}

\input{sec/4_exp}
\input{sec/5_conclusion}
\clearpage
{
    \small
    \bibliographystyle{ieeenat_fullname}
    \bibliography{main}
}

\input{sec/X_suppl}

\end{document}

%% file: preamble.tex
\newcommand{\weitai}[1]{%
  {\noindent\small\textcolor{red}{weitai: the above content has been refined. You can start to review.}}%
}

\usepackage{microtype}

\renewcommand{\paragraph}[1]{\vspace{.5em}\noindent\textbf{#1.}}

\usepackage{xspace}

\usepackage{soul}
\setuldepth{foobar}

\usepackage{graphicx}
\usepackage{amsmath}
\usepackage{amssymb}
\usepackage{booktabs}
\usepackage{multirow}

\usepackage{multicol} 
\usepackage{graphicx}

\usepackage{pifont}
\usepackage{enumitem}
\usepackage{wrapfig}
\usepackage{colortbl}
\usepackage{algorithm}
\usepackage{algorithmicx}
\usepackage{algpseudocode}
\usepackage{xcolor}  
\usepackage{varwidth}
\usepackage{arydshln}
\definecolor{mygreen}{RGB}{0,100,0}

%% file: sec/0_abstract.tex
\begin{abstract}

Current visual grounding models are either based on a Multimodal Large Language Model (MLLM) that performs auto-regressive decoding, which is slow and risks hallucinations, or on re-aligning an LLM with vision features to learn new special or object tokens for grounding, which may undermine the LLM’s pretrained reasoning ability.
In contrast,
we propose \textbf{VGent}, a modular encoder–decoder architecture that explicitly disentangles high-level reasoning and low-level bounding box prediction. 
Specifically, a frozen MLLM serves as the encoder to provide untouched powerful reasoning capabilities, 
while a decoder takes high-quality boxes proposed by detectors as queries and selects target box(es) via cross-attending on encoder's hidden states.
This design fully leverages advances in both object detection and MLLM, avoids the pitfalls of auto-regressive decoding, and enables fast inference.
Moreover, it supports modular upgrades of both the encoder and decoder to benefit the whole system: we introduce 
(i) \textbf{QuadThinker}, an RL-based training paradigm for enhancing multi-target reasoning ability of the encoder; 
(ii) \textbf{mask-aware label} for resolving detection–segmentation ambiguity; and 
(iii) \textbf{global target recognition} to improve the recognition of all the targets which benefits the selection among augmented proposals. 
Experiments on multi-target visual grounding benchmarks show that VGent achieves a new state-of-the-art with {\bf +20.6\%} F1 improvement over prior methods, and further boosts gIoU by {\bf +8.2\%} and cIoU by {\bf +5.8\%} under visual reference challenges, while maintaining constant, fast inference latency.

\end{abstract}

%% file: sec/1_intro.tex
\section{Introduction}
\label{sec:intro}

\begin{figure}[t]
    \centering
    \includegraphics[width=\linewidth]{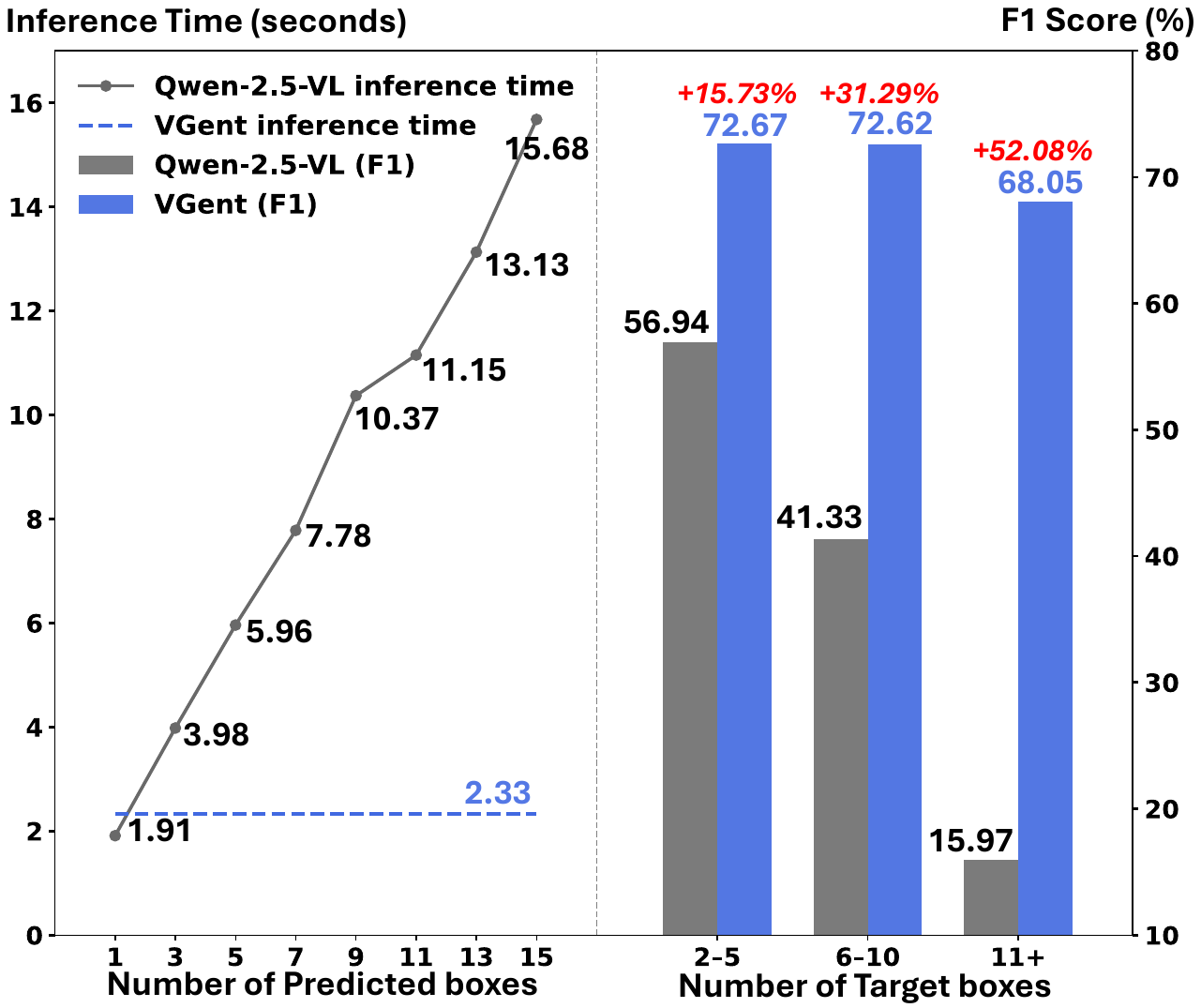}
    \vspace{-10pt}
    \caption{
\textbf{Comparison of inference speed and performance.} Auto-regressive MLLMs show linearly increasing inference time with more predicted boxes and struggle in multi-target scenarios. In contrast, VGent’s modular design enables parallel inference with constant, fast latency and superior performance, even when the number of targets grows.}
\vspace{-10pt}
    \label{fig:overview}
\end{figure}

Visual grounding~\cite{refcoco, refcocog, segvg, attbalance, actress} is a fundamental multimodal fine-grained capability, which aims to localize the referred target(s) in an image given a natural language description.
It enables human–AI interaction in real-world applications~\cite{infantagent-next,intent3d,robin3d} and serves as a crucial component for enhancing multimodal reasoning systems~\cite{gan2017vqs, qwen2.5vl, internvl3.5}.

In the era of MLLMs, many approaches leverage the pretrained reasoning capabilities of (M)LLMs and fine-tune them for grounding tasks. 
We categorize existing methods into two types:
(1) \textit{Native-token}, which follows the MLLM’s original vocabulary and decoding paradigm to generate box coordinates~\cite{shikra, kosmos2, ferret, ferret2, lmmdet, llava1.5, qwen2.5vl, expvg, guirlvg, segzero, visionreasoner} or text-as-mask~\cite{text4seg} token by token, and
(2) \textit{New-token}, which supervisedly fine-tunes the LLM space to align newly introduced special or object tokens outside the pretrained vocabulary~\cite{groma, chatscene, robin3d, lisa, pixellm, visionllm2, glamm, omgllava, lumen, intent3d}, which are decoded to the location of target.
However, both strategies have notable limitations. 
\textit{Native-token} methods are inherently slow, as each generated token must pass through the entire transformer stack, causing inference time to grow linearly with the number of targets.
They also risk hallucinations~\cite{pope, visualhallucination, hallucination}, such as prematurely stopping before enumerating all target objects or entering endless generation loops in dense-object scenes~\cite{qwen3}.
Their inefficiency and 
instability
become more evident in multi-target scenarios, as demonstrated in \cref{fig:overview}. 
\textit{New-token} methods, on the other hand, 
require collecting large-scale new datasets and performing extensive fine-tuning on a LLM to build a MLLM with the newly introduced tokens,
thereby forgoing the use of available advanced open-source MLLMs~\cite{qwen2, qwen2.5vl}
and inevitably disrupting the general reasoning capabilities of the LLM backbone acquired from pretraining~\cite{llama2, vicuna}.

These challenges highlight a fundamental conflict: forcing a single, monolithic model to excel at both abstract semantic reasoning and precise, low-level localization inevitably leads to trade-offs, degrading both efficiency and reasoning fidelity. We argue that these two capabilities are distinct and best handled by specialized components.
Motivated by this observation, we propose \textbf{VGent}, a modular encoder–decoder design that decouples high-level multimodal reasoning and low-level prediction using off-the-shelf detectors.
Our key insight is that the strengths of MLLMs and detectors are complementary: \textit{MLLMs excel at reasoning and semantic alignment, whereas detectors provide efficient and accurate localization.}
Specifically, first,
VGent's encoder is a frozen, pretrained MLLM that provides untouched reasoning abilities to interpret the image and recognize targets suggested by the language. We leverage its internal reasoning signals encoded in the hidden states.
Second,
high-quality boxes are proposed by off-the-shelf detectors. 
Third,
a decoder takes these proposals as queries and cross-attends to the encoder’s hidden states to determine which proposals correspond to the target(s).
This design fully exploits the 
high recall and reliable objectness of modern detectors
while preserving the strong reasoning capabilities of the MLLM. 
Since VGent avoids auto-regressive decoding during inference, 
we simultaneously achieve significant improvements in both inference efficiency and performance in multi-target scenarios, as shown in \cref{fig:overview}.

Additionally, the modular design of VGent enables component-wise enhancements for further performance gains:
(a) we introduce \textbf{QuadThinker}, an RL-based training paradigm tailored to incentive the encoder’s multi-target reasoning capabilities; 
(b) we propose a \textbf{mask-aware label} scheme to resolve the inherent ambiguity between detection (which focuses on a one-to-one mapping between targets and predictions) and segmentation (which focuses on recalling all pixels belonging to the target group); 
and (c) we introduce a \textbf{global target recognition} module to enhance the decoder’s ability to recognize targets globally and benefit the selection of proposals when they are augmented.

Experiments on the multi-target grounding benchmark (MaskGroups-HQ) show that VGent surpasses the previous state-of-the-art method by {\bf +20.58\%} F1.
It also improves gIoU by {\bf +8.22\%} and cIoU by {\bf +5.83\%} in the visual reference challenge, demonstrating strong reasoning over fine-grained visual prompts.
On traditional single-target grounding tasks (RefCOCO, RefCOCO+, RefCOCOg), VGent attains an average accuracy of {\bf 90.1\%}, outperforming much larger models such as InternVL3.5-20B and 38B, and improving its backbone, Qwen2.5-VL-7B, by {\bf +3.5\%}.

In sum, we make the following contributions:
(i) We propose VGent, a modular encoder–decoder framework that disentangles high-level reasoning and low-level prediction.
(ii) We introduce several modular upgrades to enhance the encoder’s reasoning capacity and the decoder’s proposal selection capability.
(iii) Extensive experiments demonstrate that VGent achieves both high efficiency and effectiveness.

%% file: sec/2_related.tex
\section{Related Work}
\label{sec:related}

\subsection{Visual Grounding and its variants}
\textit{Referring Expression Comprehension} (REC)~\cite{transvg, vltvg, segvg, attbalance} is the vanilla form of visual grounding. 
Given an image and a referential sentence that typically describes the category, attribute, or positional information of a target object, the goal is to localize the referred object by predicting its box. 
\textit{Referring Expression Segmentation} (RES)~\cite{refcoco, refcocog} extends REC to segmentation, requiring the model to predict precise pixel-level masks. 
It remains a single-target task and mask annotations in the benchmarks~\cite{refcoco, refcocog} may contain biases~\cite{sam3}.  
\textit{Generalized Referring Expression Segmentation} (GRES)~\cite{gref} further broadens RES by allowing expressions to refer to an arbitrary number of target objects.
Although more challenging, GRES still partially inherits the single-target split from RES.
Most recently, \textit{Omnimodal Referring Expression Segmentation} (ORES)~\cite{ores} generalizes RES to multi-target scenarios over diverse image domains and entities, using high-resolution images from EntitySeg~\cite{qi2022high}. 
It introduces visual references in the queries, creating fine-grained challenges for grounding multiple targets.
This makes ORES particularly suitable as a benchmark for evaluating multi-target visual grounding.

\begin{figure*}[th]
    \centering
    \includegraphics[width=0.9\linewidth]{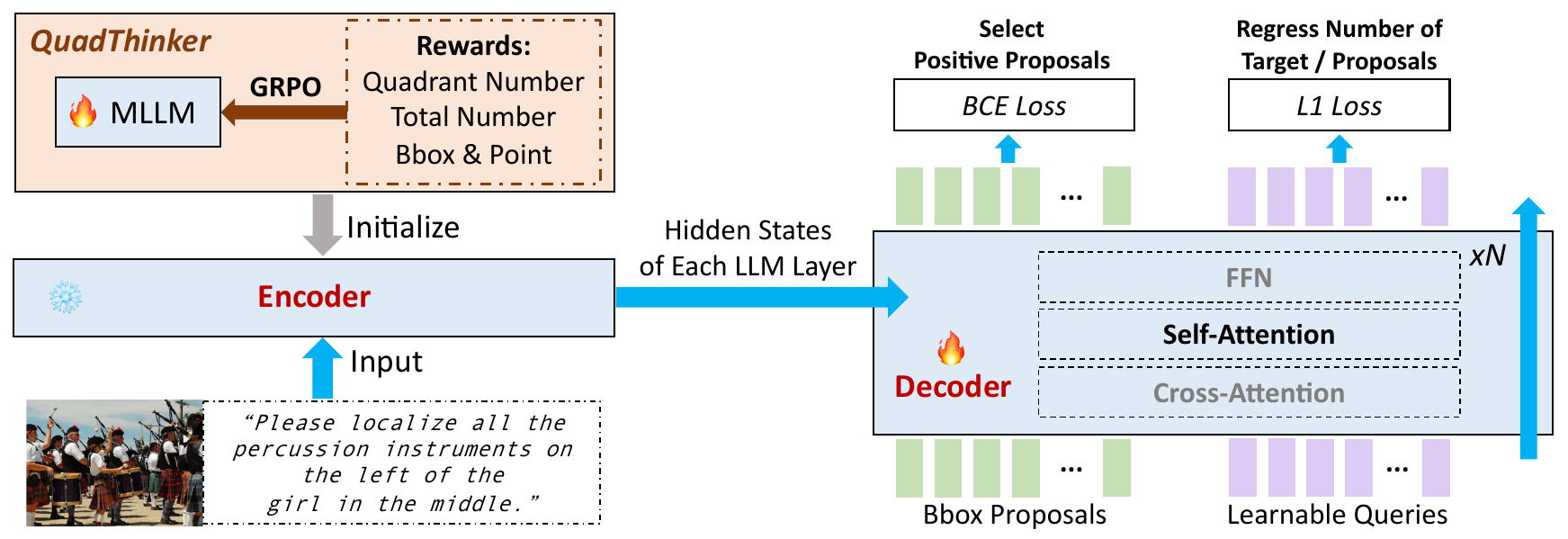}
    \vspace{-5pt}
    \caption{
\textbf{Overview of the VGent framework.}
VGent adopts a modular encoder–decoder architecture that explicitly separates high-level multimodal reasoning from low-level box prediction.
The \textbf{encoder} (left) is a frozen pretrained MLLM that processes image–text inputs jointly and stores multimodal hidden states from all transformer layers.
The \textbf{decoder} (right), initialized from the encoder’s LLM layers, takes box proposals from off-the-shelf detectors as queries and performs cross-attention with the encoder’s hidden states to select target box(es).
A self-attention layer enables interaction among proposals, while layer-wise initialization ensures reasoning–prediction alignment.
The output box queries predict object presence. 
We further involve learnable queries in the decoder for auxiliary numerical prediction. 
}
\vspace{-3pt}
    \label{fig:method}
\end{figure*}

\subsection{MLLM for Visual Grounding}
We categorize the existing MLLM visual grounding methods into two types: \textit{Native-token} and \textit{New-token}.
\textit{Native-token} represents a line of works that directly leverage the original vocabulary of MLLMs to auto-regressively output box coordinates (e.g.  LLaVA-1.5~\cite{llava1.5}, Qwen2.5-VL~\cite{qwen2.5vl}, Shikra~\cite{shikra}, KOSMOS-2~\cite{kosmos2}, Ferretv2~\cite{ferret2}, and LMM-Det~\cite{lmmdet}) or text-as-mask (Text4Seg~\cite{text4seg}) as tokens. While this paradigm aligns well with MLLM pretraining objectives, it is inherently slow and prone to hallucinations as the number of targets increases~\cite{qwen3}.
\textit{New-token} refers to another line of approaches that introduce newly added tokens outside the original LLM vocabulary to represent object entities. Some methods introduce new tokens corresponding to object identifiers and decode them auto-regressively (e.g., Groma~\cite{groma}, Chat-Scene~\cite{chatscene}, Robin3D~\cite{robin3d}) to indicate the referred objects. Others append object features to the sequence and perform classification over each object feature (e.g., RAS~\cite{ores}).
In addition,
several works compress target information into a new vocabulary token (e.g., “\texttt{[Det]}” or “\texttt{[Seg]}”), which is subsequently decoded into a box or mask by a downstream module (e.g., LISA~\cite{lisa}, PixelLM~\cite{pixellm}, VisionLLMv2~\cite{visionllm2}, GLaMM~\cite{glamm}, OMG-LLaVA~\cite{omgllava}).

%% file: sec/3_method.tex
\section{Methodology}
\label{sec:method}
We first present the overall VGent framework (Sec.~\ref{sec:vgent_framework}), which consists of an encoder and a decoder along with a detector.
We then describe three modular enhancements—\textit{QuadThinker} for the encoder, \textit{mask-aware label} for the decoder, and \textit{global target recognition} for the detector and decoder—to further enhance the performance (Sec.~\ref{sec:vgent_modular}).

\subsection{VGent Framework}
\label{sec:vgent_framework}

VGent is a modular encoder–decoder framework designed to explicitly separate high-level multimodal reasoning from low-level (pixel-level) localization.

\paragraph{Encoder}
As shown on the left of \cref{fig:method}, the encoder is initialized from a pretrained MLLM. To ensure it possesses strong multi-target reasoning capabilities, we first enhance the base MLLM using our QuadThinker paradigm (detailed in \cref{quadthinker}). The resulting model
is then frozen and used as the encoder, preserving its multi-target capabilities.
Given an image and a text, the encoder MLLM projects vision features from the vision encoder into the LLM space and concatenates both visual and textual tokens to form a multimodal sequence.
This sequence passes through all transformer layers of the LLM, and we store the hidden states from each layer, which capture information at different levels—from basic object identity and counting in shallow layers to abstract semantic clues in deeper ones~\cite{vfl, fan2024not, oota2023joint, rahimi2025explanations}.

\paragraph{Decoder}
As shown on the right of \cref{fig:method}, the decoder is a transformer initialized from the LLM part of the encoder. 
It takes the box proposal from off-the-shelf detectors as the queries, while its keys and values are taken from the encoder’s hidden states.
Specifically, the image is first processed by a detector to generate $N$ proposals $p \in \mathbb{R}^{N \times 4}$. These proposals are projected through an MLP into the LLM space to produce query embeddings $q \in \mathbb{R}^{N \times C}$, where $C$ is the LLM’s hidden dimension. 
In each decoder layer $i$, the queries come from the output of its previous layer, and the key–value pairs are set to the output of $(i-1)$-th layer of the encoder LLM. 
Since each decoder layer is initialized from its corresponding encoder layer, this layer-wise alignment enables the decoder to effectively interpret the reasoning signals encoded in the key–value pairs.
Within the decoder layer, the cross-attention module is used to initialize a subsequent self-attention module, which allows proposal queries to exchange information and jointly identify targets—particularly when combined with the {global target recognition} module in \cref{sec:global_target}.
A feed-forward network follows to produce the layer output.
Finally, an MLP head processes the output queries from the last layer to predict whether each proposal corresponds to a target object.
Binary cross-entropy loss is used for supervision, where proposals 
exceeding a certain IoU threshold with any ground-truth box are treated as positive and others as negative.
Auxiliary losses for learnable queries are elaborated in \cref{sec:global_target}. 

\subsection{Modular Enhancements}
\label{sec:vgent_modular}

VGent’s modular design enables targeted improvements to the encoder and decoder to further boost performance. We introduce three key enhancements: QuadThinker for strengthening encoder reasoning, mask-aware label for bridging detection–segmentation gaps, and global target recognition for improving proposal selection.

\begin{figure}[t]
    \centering
    \includegraphics[width=\linewidth]{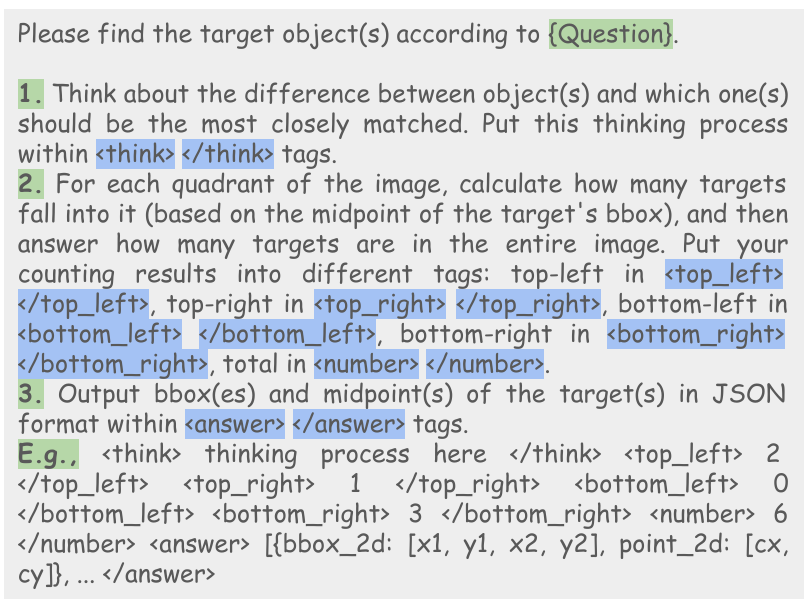}
    \vspace{-14pt}
    \caption{
Prompt for GRPO training of \textbf{QuadThinker}. Key components of the prompt are highlighted in \textcolor{mygreen}{green}, while specific instructions used for verifiable rewards are highlighted in \textcolor{blue}{blue}.
}
\label{fig:prompt}
\end{figure}

\subsubsection{QuadThinker\\--\textit{Reinforcing Multi-target Reasoning}}\label{quadthinker}
We observe that pretrained MLLMs degrade notably as the number of target objects increases (\cref{fig:overview}), even though their pretraining data contains multi-object scenes~\cite{qwen2.5vl}. This suggests that multi-target grounding remains the main bottleneck. To address this, we introduce \textbf{QuadThinker}, an RL-based fine-tuning paradigm built on GRPO~\cite{grpo} to enhance the encoder’s multi-target reasoning ability. 
The key idea is to design prompts and verifiable reward functions that explicitly guide the model to perform region-to-global, step-by-step reasoning, thereby reducing hallucinations and improving its ability to handle multi-target scenarios.
Specifically, given the prompt in \cref{fig:prompt}, 
the model needs to first recognize the targets within each image quadrant by predicting the target counts, then summarize the overall number of targets. After this instance-level recognition, the model is further required to predict the boxes and center points of each target. We introduce a \textit{format reward function}, which evaluates whether the model’s response adheres to the required step-by-step reasoning format to contain all necessary tags. Additionally, we propose an \textit{accuracy reward function}, which measures how well the predicted quadrant-wise counts, total counts, and box/point coordinates align with the ground truth. The detailed procedure is in Algorithm~\ref{alg:quadthinker_reward}.

\begin{algorithm}[t]
\caption{Reward Computation in \textbf{QuadThinker}}
\label{alg:quadthinker_reward}
\centering
\begin{minipage}{0.9\textwidth}  
{\small  
\begin{algorithmic}[1]
\Require Prediction $P$, Ground truth $G$, image dimensions 
\Ensure Total reward $R_{\text{total}}$
\State Initialize $R_{\text{total}} \gets 0$
\Statex \textcolor{blue}{// --- Format Reward Function---}
\If{$P$ contains all required tags}
    \State $R_{\text{total}} \gets R_{\text{total}} + 1.0$
\EndIf
\If{All count tags contain valid integers}
    \State $R_{\text{total}} \gets R_{\text{total}} + 1.0$
\EndIf
\If{\texttt{answer} tag contains valid JSON}
    \State $R_{\text{total}} \gets R_{\text{total}} + 2.0$
\EndIf
\Statex \textcolor{blue}{// --- Accuracy Reward Function---}
\State Parse $G$ for boxes, centroids, and counts
\State Parse $P$ for boxes, centroids, and counts
\If{All counts match}
    \State $R_{\text{total}} \gets R_{\text{total}} + 1.0$
\EndIf
\State Compute reward indicators: $R_{\text{IoU}} = \mathbf{1}[\text{IoU} > 0.5]$,  
\Statex \quad $R_{\text{L1}} = \mathbf{1}[\text{L1} < 10]$, 
$R_{\text{point}} = \mathbf{1}[\text{dist} < 30]$
\State Construct cost matrix: $C = 3.0 - (R_{\text{IoU}} + R_{\text{L1}} + R_{\text{point}})$
\State Apply hungarian matching on $C$ to compute $R_{\text{det}}$
\State $R_{\text{total}} \gets R_{\text{total}} + R_{\text{det}}$
\State \Return $R_{\text{total}}$
\end{algorithmic}
}
\end{minipage}
\end{algorithm}

\begin{figure*}[th]
    \centering
    \includegraphics[width=0.85\linewidth]{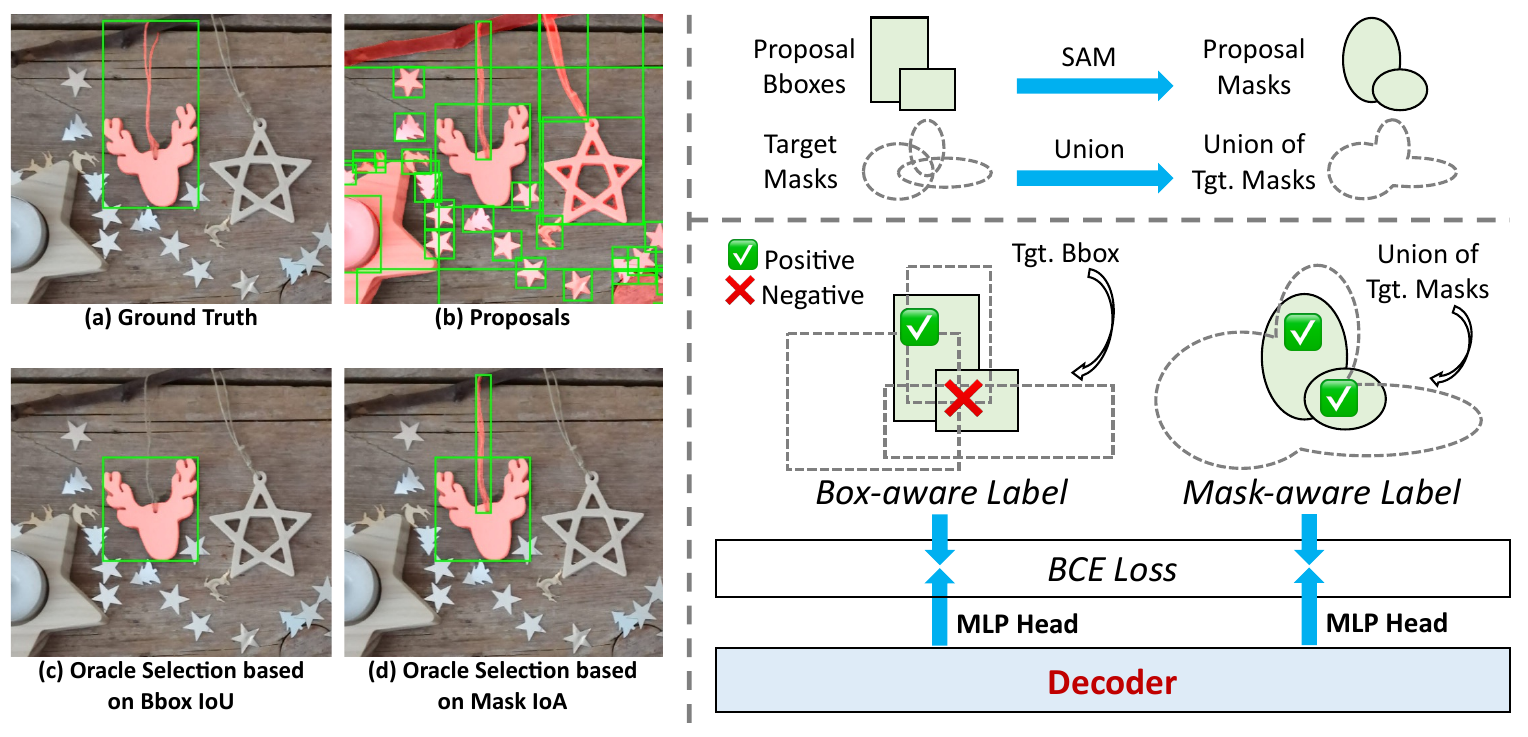}
    \vspace{-4pt}
    \caption{
Comparison between IoU-based and IoA-based labeling and the design of the proposed Mask-aware Label.
(Left) Example from the MaskGroups-HQ dataset.
(a) The ground-truth mask includes both the deer decoration and its attached string.
(b) Detector proposals treat them as two separate objects.
(c) Even with oracle selection (Hungarian matching with IoU $> 0.5$), small yet valid regions (e.g., the string) are missed.
(d) The proposed IoA-based Mask-aware Label captures these fine-grained regions (i.e., the string) by normalizing intersection over proposal's area.
(Right) Overview of the Mask-aware Label mechanism.
(Top-Right) Proposal masks are obtained by prompting SAM; all ground-truth masks are unified into one mask to compute IoA for label assignment.
(Bottom-Right) Two MLP heads predict labels separately for detection (box-aware) and segmentation (mask-aware) tasks, respectively.
}
\vspace{-10pt}
    \label{fig:masklabel}
\end{figure*}

\subsubsection{Mask-aware Label\\--\textit{Bridging Detection and Segmentation}}
\label{sec:masklabel}

We observe a significant gap between detection and segmentation tasks, mainly caused by annotation ambiguity and the inconsistent granularity of proposals. Using the MaskGroups-HQ dataset~\cite{ores} as an example—which involves multiple targets—we convert each ground-truth mask into a bounding box to analyze the selection behavior.
As illustrated in \cref{fig:masklabel}-left-(a), the ground-truth annotation of the deer head decoration includes both the decorative head and the string attached to it. However, in the corresponding 
box
proposals shown in \cref{fig:masklabel}-left-(b), whose masks are obtained via prompt-based SAM, the detector does not consider the string and decoration as a unified object. Instead, it generates two separate 
boxes: one covering the main decoration body and another covering the string.
Detection typically optimizes one-to-one bipartite matching. Therefore, even with oracle selection (Hungarian matching followed by filtering proposals with Intersection-over-Union (IoU) $> 0.5$), as shown in \cref{fig:masklabel}-left-(c), the string proposal cannot be selected—leading to missed regions. In contrast, segmentation focuses on retrieving all foreground pixels, meaning that small or fragmented proposals that partially overlap with the annotated region should ideally be retained.

To address this discrepancy, we introduce the \textbf{Mask-aware Label}, which uses a new metric—Intersection-over-Area (IoA)—for label assignment during training. Specifically, as shown in \cref{fig:masklabel}-top-right, we get the mask of each proposal by prompting SAM~\cite{sam2} and unify all the ground truth masks as one mask.
We then compute the intersection between each SAM-generated proposal mask and the ground-truth union mask, divided by the area of the proposal mask. As illustrated in \cref{fig:masklabel}-left-(d), this normalization by proposal's area enables the labeling to identify small but valid proposals (e.g., the string). When the IoA exceeds 0.6, the proposal is labeled as positive; otherwise, it is labeled as negative.
We refer to the conventional IoU-based labeling as box-aware label. As shown in \cref{fig:masklabel}-top-down, the model employs two separate MLP heads to predict the two types of labels: the box-aware head for detection tasks, and the mask-aware head for segmentation tasks.

\subsubsection{Global Target Recognition\\--\textit{Improving Proposal Selection}}
\label{sec:global_target}

\begin{figure}[t]
    \centering
    \includegraphics[width=\linewidth]{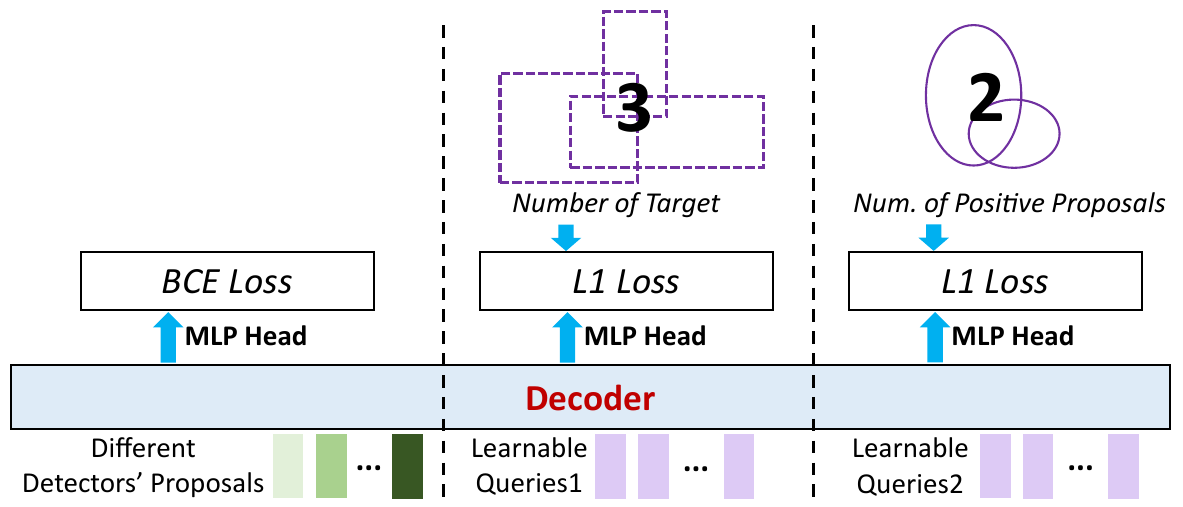}
\vspace{-10pt}
    \caption{
Illustration of the proposed Global Target Recognition mechanism.
Proposals from multiple detectors are aggregated into a unified query set to improve recall.
A small set of learnable queries is concatenated with the proposal queries before entering the decoder.
Half of these learnable queries predict the total number of targets, while the other half estimate the number of positive proposals based on the mask-aware label.
Through self-attention, the learnable queries inject global target information into each proposal, enabling more holistic and accurate proposal selection.
}
\vspace{-10pt}
    \label{fig:numQ}
\end{figure}

To further strengthen the model’s proposal selection capability, we introduce \textbf{Global Target Recognition}, which improves each proposal’s global awareness of all targets, particularly under proposal augmentation.
As illustrated in \cref{fig:numQ}, we aggregate proposals generated from multiple detectors and concatenate them into a unified set of proposal queries, which increases the recall of target objects.
In addition, we introduce a small set of \textit{learnable queries}, which are concatenated with the proposal queries to form the final input to the decoder.
During decoding, half of these learnable queries are trained to predict the total number of target objects, while the other half are optimized to estimate the number of positive proposals based on the mask-aware label. The ground truths are normalized by 1000 and we use L1 loss as the objective function.
These learnable queries thus encode global target information and interact with proposal queries through the decoder’s self-attention layers.
This design allows global cues to be propagated to each proposal, enhancing its holistic understanding of the target group and leading to more accurate proposal selection.

\begin{table*}[th]
    \centering
    \vspace{-2mm}
    \caption{
\textbf{Results on Omnimodal Referring Expression Segmentation (ORES).} 
ORES provides high-quality human-annotated visual grounding data covering both single- and multi-target expressions, including a referential split (w/ \texttt{<mask-ref>}) where queries involve spatial references. 
VGent achieves new state-of-the-art performance across all metrics, showing consistent gains over strong baselines such as RAS$_\text{13B}$ and Qwen3-VL-30B, and demonstrating robust generalization under referential conditions.
}
    \begin{tabular}{l c c c c c c c c c}
        \toprule
        \multirow{3}{*}{{Model}} & \multicolumn{3}{c}{w/o \texttt{\textless mask-ref\textgreater}} & \multicolumn{3}{c}{w/ \texttt{\textless mask-ref\textgreater}} & \multicolumn{3}{c}{Overall} \\
        \cmidrule(lr){2-4}\cmidrule(lr){5-7}\cmidrule(lr){8-10}
         & F1 & gIoU & cIoU & F1 & gIoU & cIoU & F1 & gIoU & cIoU \\
        \midrule
        ReLA~\cite{liu2023gres} & - & 34.93 & 43.22 & - & - & - & - & - & - \\
        PSALM$_\text{1.3B}$~\cite{zhang2024psalm} & - & 36.92 & 37.33 & - & - & - & - & - & - \\
        GSVA$_\text{13B}$~\cite{xia2024gsva} & - & 41.98 & 49.55 & - & - & - & - & - & - \\
        RAS$_\text{13B}$~\cite{ores} & 51.65 & 66.71 & 74.59 & 48.80 & 58.72 & 68.77 & 50.89 & 64.77 & 73.13 \\
        Qwen3-VL-30B-A3B-Instruct~\cite{qwen3} & 60.50 & 64.79 & 64.81 & 34.98 & 41.40  & 39.34  & 53.23  & 58.76  & 57.61  \\
        \midrule
        VGent (Ours) & \bf 71.85 &  \bf 68.89 & \bf 75.50 & \bf 70.45 & \bf 66.94 & \bf 74.60 & \bf 71.47 & \bf 68.42 & \bf 75.28 \\
        \bottomrule
    \end{tabular}
    \vspace{-2mm}
    \label{tab:ores}
\end{table*}

%% file: sec/4_exp.tex
\section{Experiments}

For the main experiments, we evaluate the model on the most recent multi-target visual grounding benchmark, Omnimodal Referring Expression Segmentation (ORES). 
We follow the previous practise~\cite{ores} to report gIoU and cIoU to measure segmentation performance. 
To get segmentation masks, we prompt SAM~\cite{sam2} by our predicted boxes.
However, both metrics are sensitive to large areas and cIoU is particularly affected, without differentiation on different instances.
To mitigate this bias and better reflect multi-target grounding capability, we also report the F1 score, which captures the precision–recall balance of instance detection.
For single-target visual grounding, referring expression segmentation benchmarks~\cite{refcoco, refcocog} exhibit significant mask annotation bias, where the language part is insufficient to uniquely identify the ground-truth mask, as confirmed by our findings and prior studies (e.g., SAM3~\cite{sam3}). Therefore, we adopt the detection setting—Referring Expression Comprehension (REC)—for evaluation. Additional experimental results on other benchmarks are provided in the Supplementary.

\begin{table*}[h] 
\centering 
\caption{\textbf{Results on referring expression comprehension (REC).}
We evaluate single-target visual grounding on RefCOCO, RefCOCO+, and RefCOCOg benchmarks~\cite{refcoco,refcocog}.
VGent achieves competitive or superior accuracy across datasets, outperforming strong MLLMs such as Qwen2.5-VL and InternVL3 series, demonstrating robust reasoning and localization abilities in single-target grounding. 
}
\begin{tabular}{l c c c c c c c c c}
\toprule
\multirow{3}{*}{{Model}}           & \multicolumn{3}{c}{{{RefCOCO}}} & \multicolumn{3}{c}{{{RefCOCO+}}} & \multicolumn{2}{c}{{{RefCOCOg}}} & \multirow{2}{*}{{{Avg.}}} \\
\cmidrule(lr){2-4}\cmidrule(lr){5-7}\cmidrule(lr){8-9}
                                        & {val}    & {test-A}    & {test-B}    & {val}    & {test-A}    & {test-B}     & {val}     & {test}                 &         \\
\midrule
Gemini2.5-Pro-thinking~\cite{gemini} & -  & -  & - & -  & -  & - & -  & -  & 74.6    \\
SegVG~\cite{segvg} & 86.8 & 89.5  & 83.1 & 77.2 & 82.6 & 67.6 & 78.4 & 77.4 & 80.3 \\
AttBalance~\cite{attbalance} & 87.3  & 89.6 & 83.9 & 77.5 & 82.0 & 68.6 & 79.86 & 79.63 & 81.1 \\
ExpVG~\cite{expvg} & 87.4 & 91.7 & 81.5  & 80.3 & 86.9 & 71.1  & 81.3 & 81.4 & 82.7 \\
Grounding-DINO-L~\citep{grounding_dino} & 90.6   & 93.2      & 88.2      & 82.8   & 89.0      & 75.9       & 86.1    & 87.0                 & 86.6    \\
UNINEXT-H~\citep{yan2023universal}               & 92.6   & 94.3      & 91.5      & 85.2   & 89.6      & 79.8       & 88.7    & 89.4                 & 88.9    \\
ONE-PEACE~\citep{one-peace}             & 92.6   & 94.2      & 89.3      & 88.8   & 92.2      & 83.2       & 89.2    & 89.3                 & 89.8    \\
Ferret-v2-13B~\citep{ferret2}          & 92.6   & 95.0      & 88.9      & 87.4   & 92.1      & 81.4       & 89.4    & 90.0                 & 89.6    \\
Qwen2-VL-7B~\cite{qwen2}      & 91.7   & 93.6      & 87.3      & 85.8   & 90.5      & 79.5       & 87.3    & 87.8                 & 87.9    \\
Qwen2.5-VL-7B~\cite{qwen2.5vl}       & 90.0   & 92.5      & 85.4      & 84.2   & 89.1      & 76.9       & 87.2    & 87.2                 & 86.6    \\
InternVL3-8B~\cite{internvl3}                            & 92.5   & 94.6      & 88.0      & 88.2   & 92.5      & 81.8       & 89.6    & 90.0                 & 89.6    \\
InternVL3-9B~\cite{internvl3}                            & 91.8  & 93.2     & 86.6     & 86.4  & 91.0     & 79.9      & 88.0   & 88.5                & 88.2    \\  
InternVL3-14B~\cite{internvl3}                           & 92.0  & 94.4     & 87.8     & 87.4  & 92.1     & 81.5      & 88.6   & 89.3                 & 89.1    \\  
InternVL3.5-8B~\cite{internvl3.5} & 92.4 & 94.7 & 88.7 & 87.9 & 92.4 & 82.4 & 89.6 & 89.4 & 89.7  \\
InternVL3.5-20B-A4B~\cite{internvl3.5} & 91.9 & 94.1 & 88.8 & 87.6 & 92.0 & 82.7 & 89.1 & 90.0 & 89.5  \\
InternVL3.5-38B~\cite{internvl3.5} & 90.3 & 91.8 & 89.0 & 87.5 & 90.0 & 84.7 & 89.7 & 89.9 & 89.1  \\
\midrule
VGent (Ours) & 92.4  & 94.7  & 89.8 & 88.1  & 92.2  & 83.3 & 90.4  & 90.1  & \bf 90.1   \\
\bottomrule
\end{tabular}
\label{tab:refcoco}
\end{table*}

\subsection{Implementation}

For ORES evaluation, we train our model on a combination of Object365~\cite{objects365}, MaskGroups-2M~\cite{ores}, and MaskGroups-HQ~\cite{ores} training sets. For REC evaluation, we follow RAS~\cite{ores} to fine-tune on the training sets of RefCOCO, RefCOCO+, and RefCOCOg~\cite{refcoco, refcocog, refcocog}.
The BCE loss is weighted by 1, and the L1 loss is weighted by 10. The learning rate is set to 2e-5 and linearly decayed.
For QuadThinker in the final performances, which is used to initialize the encoder of VGent, we perform GRPO training for one epoch based on Qwen2.5-VL-7B~\cite{qwen2.5vl} using the MaskGroups-HQ~\cite{ores} training set and VisionReasoner-7K~\cite{visionreasoner}, with a batch size of 16 and a learning rate of 1e-6. VGent has around 15.7B parameters.
Additional details are provided in the Supplementary.

\subsection{Quantitative Results}

\subsubsection{Multi-target Visual Grounding}

ORES (MaskGroups-HQ)~\cite{ores} is a recent high-quality visual grounding dataset that contains both single- and multi-target expressions. Each sample is human-annotated with strict quality control, and the language queries support referential masks in the expressions, represented by the \texttt{<mask-ref>} split. We convert these referential masks into box coordinates so that they can be incorporated into the language representation. Details for this are provided in the Supplementary.
Unlike COCO-based benchmarks, ORES features higher-resolution images and richer entity-level annotations, making it a more challenging testbed for visual grounding.
We also evaluate the latest MLLM as of the time of writing, Qwen3-VL-30B-A3B-Instruct, on this benchmark. 
Its segmentation results are obtained using SAM-based prompting, consistent with our setup. Details are in the Supplementary.

As shown in \cref{tab:ores}, even Qwen3-VL (a model with larger scale than ours) exhibits
suboptimal
performance in the multi-target setting, despite its major improvements in multi-object detection tasks~\cite{qwen3}.
This observation suggests that while single-target visual grounding has become nearly saturated, \textit{multi-target grounding remains a major bottleneck in visual grounding.}
In contrast, VGent achieves new state-of-the-art results across all metrics and both splits, surpassing the previous strong baseline RAS$_\text{13B}$~\cite{ores}.
Specifically, VGent brings a substantial improvement of {\bf +20.58\%} F1 overall, including {\bf +20.2\%} on the w/o \texttt{<mask-ref>} split and {\bf +21.65\%} on the w/ \texttt{<mask-ref>} split. These results highlight the advantages of our modular design, which fully leverages the detector’s high recall while avoiding the MLLM’s autoregressive token-by-token generation process that often suffers from hallucinations when the number of targets increases and the output sequence becomes longer.

Notably, models generally struggle on the more challenging w/ \texttt{<mask-ref>} split which further requires reasoning on fine-grained visual references, indicating that \textit{visual grounding under visual prompts represents another key bottleneck}.
However, through the decoding design on hidden-state, VGent effectively exploits the intrinsic reasoning capability of MLLM to enhance reasoning of visual prompts.
Eventually, VGent achieves a significant improvement of {\bf +8.22\%} gIoU and {\bf +5.83\%} cIoU on  w/ \texttt{<mask-ref>} split.

In summary, VGent’s modular design fully leverages both the detector and the MLLM, enabling superior handling of complex, multi-target grounding scenarios.

\subsubsection{Single-target Visual Grounding}

To follow previous visual grounding studies, we further evaluate VGent on traditional single-target benchmarks, including RefCOCO, RefCOCO+, and RefCOCOg. 
As shown in \cref{tab:refcoco}, VGent reaches an average accuracy of {90.1\%}, surpassing previous models that are larger in size and equipped with newer backbones, such as InternVL3.5-20B and InternVL3.5-38B. 
Compared to our backbone, {Qwen2.5-VL-7B}, VGent achieves a significant improvement of \textbf{+3.5\%} on average. 
Specifically, it brings a \textbf{+4.4\%} improvement on {RefCOCO testB}, a remarkable \textbf{+6.4\%} gain on the more challenging {RefCOCO+ testB}, and a \textbf{+3.2\%} increase on {RefCOCOg val}, where language expressions are typically longer. 
These gains can be attributed to the {QuadThinker}, which enhances reasoning capability by GRPO training, and VGent’s hidden-state decoding mechanism, which effectively interprets the model’s internal reasoning process.

\begin{figure*}[t]
    \centering
    \includegraphics[width=\linewidth]{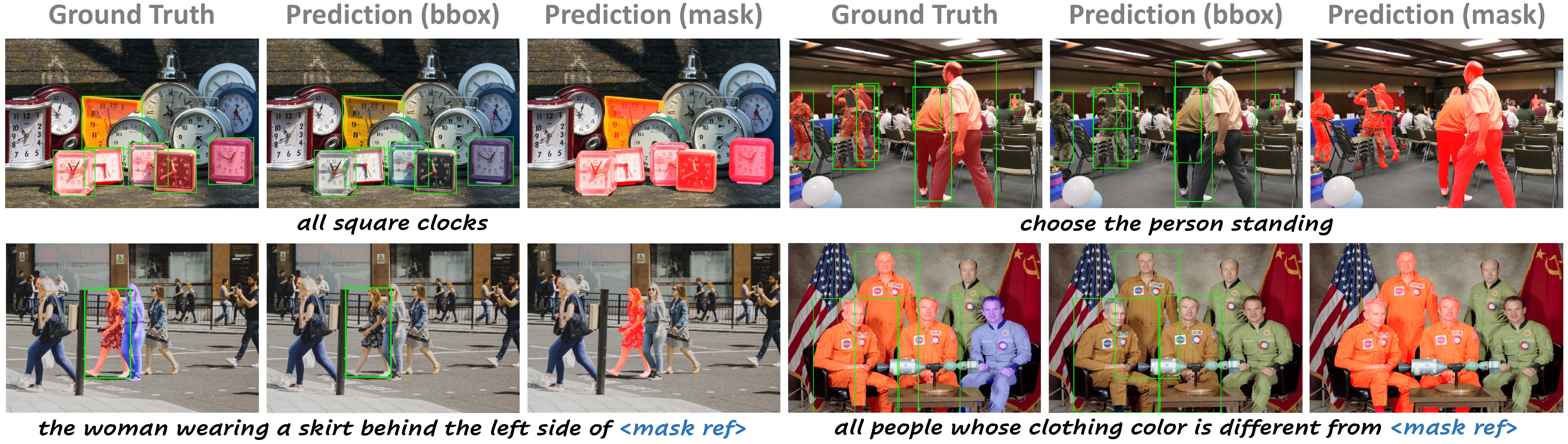}
    \vspace{-13pt}
    \caption{Visualizations of VGent's output under different challenges. Blue masks indicate visual reference regions.}
    \vspace{-6pt}
\label{fig:vis}
\end{figure*}

\begin{table}[t]
\centering
\caption{
Ablation results on {MaskGroups-HQ w/o \texttt{<mask-ref>}}.
We report F1 scores (\%) across different numbers of targets.
“Detection RL” refers to reinforcement learning with think-answer format and detection-based rewards and formats,
and “Number RL” adds the number-based reward with corresponding format.
“VGent” denotes plugging a backbone into the VGent framework.
“Full Train” indicates jointly training both the encoder and decoder.
}
\resizebox{0.48\textwidth}{!}{
\begin{tabular}{llcccc}
\toprule
{ID} & {Method} & {Total} & {2--5 Targets} & {6--10 Targets} & {11+ Targets} \\
\midrule
(1) & Qwen-2.5-VL & 45.72 & 56.94 & 41.33 & 15.97 \\
(2) & (1) + Detection RL & 54.89 & 59.30 & 56.79 & 41.43 \\
(3) & (2) + Number RL & 58.17 & 60.70 & 61.35 & 50.39 \\
\hdashline
(4) & (1) + VGent & 58.77 & 60.00 & 64.33 & 53.84 \\
(5) & (3) + VGent & \textbf{60.55} & \textbf{62.59} & \textbf{65.07} & \textbf{54.53} \\
(6) & (5) + Full Train & 45.66 & 43.76 & 53.26 & 49.39 \\
\bottomrule
\end{tabular}
}
\label{tab:ablation1}
\end{table}

\begin{table}[t]
\centering
\caption{Ablation on decoder-side enhancements on {MaskGroups-HQ} w/o \texttt{<mask-ref>}. We report F1, gIoU, and cIoU to evaluate the segmentation-oriented improvements. Both mask-aware label and global target recognition progressively strengthen VGent’s holistic reasoning and multi-detector synergy.}
\resizebox{0.44\textwidth}{!}{
\begin{tabular}{llccc}
\toprule
{ID} & {Method} & {F1} & {gIoU} & {cIoU} \\
\midrule
(7) & (5) + HQ & 69.70 & 65.02 & 65.84 \\
(8) & (7) + Mask-aware Label & 70.47 & 67.06 & 69.35 \\
(9) & (8) + Global Target Recognition & \textbf{71.60} & \textbf{69.72} & \textbf{72.78} \\
\bottomrule
\end{tabular}}
\label{tab:ablation2}
\end{table}

\subsection{Ablation Study}
\label{sec:ablation}

We conduct comprehensive ablation studies to validate the effectiveness of our proposed components. 
The experiments are divided into two major parts: 
(1) examining the reward design of {QuadThinker} and the overall modular design of {VGent} in \cref{tab:ablation1}, 
and (2) analyzing the contribution of decoder-side enhancements, including {mask-aware label} and {global target recognition} in \cref{tab:ablation2}.

\paragraph{Effect of QuadThinker and Modular VGent}
To avoid confounding factors, we adopt a stepwise ablation study. 
All the evaluations are conducted
on the {MaskGroups-HQ w/o \texttt{<mask-ref>}} split and report F1 scores across different ranges of target counts. 
We start from the {Qwen2.5-VL} backbone and progressively integrate our proposed modules.
For {QuadThinker}-related comparisons, we apply GRPO training on the VisionReasoner-7K dataset for one epoch. 
For experiments involving training VGent's decoder, 
we additionally include Object365 to provide multi-target data and train for 8K steps. 
UPN~\cite{upn} is used as the default proposal generator.

As shown in \cref{tab:ablation1}, starting from {Qwen2.5-VL} (ID (1)), adding reinforcement learning with detection-based rewards (ID (2))—including think-answer format and box / point prediction—leads to clear improvements. 
Further introducing the {number-based reward} (ID (3)), which requires the model first to predict quadrant-wise and global target counts before detection, enables explicit region-to-global, step-by-step reasoning. 
This design notably improves performance in challenging multi-target scenarios, bringing a gain of \textbf{+8.96\%} when the number of targets exceeds 11.
When integrating Qwen2.5-VL backbone into {VGent} (ID (4)), compared to the plain backbone (ID (1)), our modular design fully leverages the detector’s high recall, achieving a remarkable \textbf{+37.87\%} improvement in scenes with over 11 targets. 
Replacing the backbone with the stronger {QuadThinker} (ID (5)) further enhances the overall reasoning capability, 
demonstrating that VGent can effectively leverage improvements in the encoder in a modular manner.
Interestingly, when we jointly train VGent’s encoder and decoder (ID (6)), the performance drops significantly, despite having more trainable parameters. 
This suggests that VGent’s reasoning ability primarily stems from the {frozen encoder}; unfreezing it disrupts the pretrained reasoning skills, leading to degraded performance.

\paragraph{Effect of Decoder Enhancements}
\Cref{tab:ablation2} further investigates the decoder-side contributions, which require mask-level annotations. 
Therefore, we fine-tune 
VGent's decoder
on the 
MaskGroups-HQ
training set for 8K steps,
and additionally report gIoU and cIoU to evaluate segmentation performance.
Adding the {mask-aware label} (ID 8) consistently improves the IoU metrics by recalling proposals with high intersection-over-area (IoA). Specifically, it yields a {\bf +2.04\%} gain on gIoU and {\bf +3.51\%} on cIoU compared to ID 7.
Further introducing {global target recognition} (ID 9) provides an additional {\bf +2.66\%} improvement on gIoU and {\bf +3.43\%} on cIoU, confirming that the number-wise global information shared among proposals enhance the holistic understanding. Moreover, this demonstrates VGent’s ability to leverage multiple detectors to achieve higher recall and more comprehensive grounding.

\subsection{Qualitative Results}
\label{sec:exp}

In \cref{fig:vis}, we showcase VGent’s strong visual grounding capability across diverse and challenging scenarios.
In the first row, VGent demonstrates robust multi-target grounding performance. Both the clock and person examples contain numerous visually similar distractors and heavy occlusions. Despite this, the model correctly identifies square clocks among various clocks and the standing person among many individuals, even when the target is far from the camera with only a few visible pixels.
In the second row, VGent handles fine-grained visual references effectively. For instance, in the lower-left example, it correctly interprets the reference and distinguishes the woman on the left side is target, though both sides contain women wearing skirts. The lower-right example further combines both visual reference and multi-target challenges, and VGent successfully resolves both.

%% file: sec/5_conclusion.tex
\section{Conclusion}

We present \textbf{VGent}, a modular encoder–decoder framework for visual grounding that disentangles high-level multimodal reasoning from low-level bounding box prediction. A frozen MLLM serves as the encoder to provide strong reasoning capabilities, while a decoder selects target box(es) from high-quality proposals by cross-attending to the encoder’s hidden states. The modular design allows further enhancements, including \textbf{QuadThinker}, \textbf{mask-aware labels}, and \textbf{global target recognition}, which improve multi-target reasoning and proposal selection. Experiments on multi-target and single-target benchmarks demonstrate that VGent achieves state-of-the-art performance while maintaining fast and constant inference, highlighting our effectiveness and efficiency.

%% file: sec/X_suppl.tex
\clearpage
\setcounter{page}{1}

\twocolumn[{%
\renewcommand\twocolumn[1][]{#1}%
\maketitlesupplementary
\vspace{-20pt}
\begin{center}
    \centering
    \captionsetup{type=figure}
    \includegraphics[width=.96\textwidth]{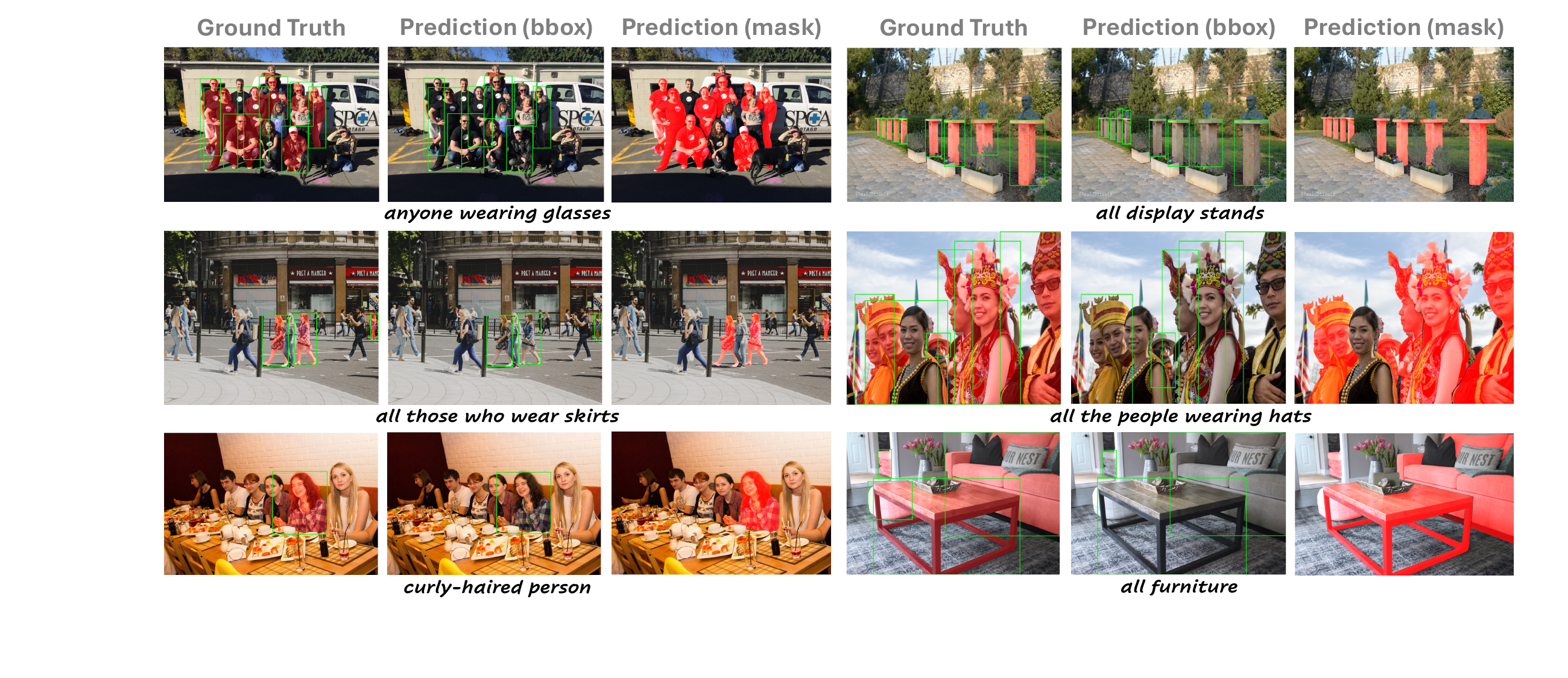}
    \captionof{figure}{Visualizations of VGent's output under single target and multiple targets challenges.}
    \label{fig:sup-vis-wo-ref}
\end{center}%
\begin{center}
    \centering
    \captionsetup{type=figure}
    \includegraphics[width=.96\textwidth]{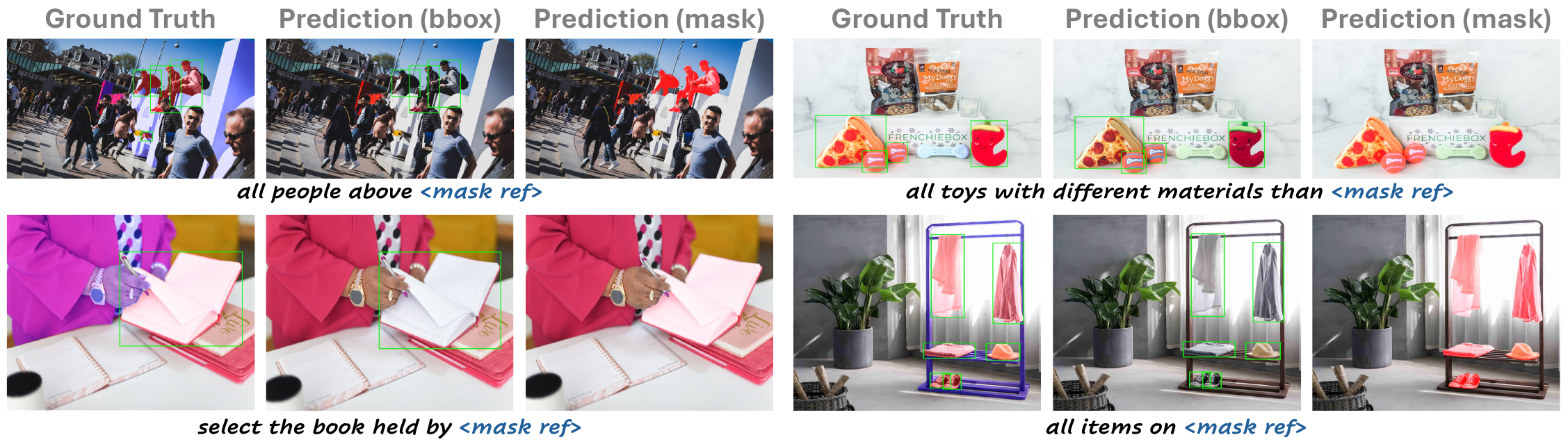}
    \captionof{figure}{Visualizations of VGent's output under visual reference challenges. Blue masks indicate visual reference regions.}
    \label{fig:sup-vis-w-ref}
\end{center}%
}]

\section{Additional Qualitative Results}
\label{sec:additional_visualizations}

In \cref{fig:sup-vis-wo-ref} and \cref{fig:sup-vis-w-ref}, we present additional qualitative examples to further illustrate the versatility and robustness of VGent across a wide range of grounding conditions, including single-target, multi-target, and visual reference–conditioned multi-target scenarios. 
These examples highlight VGent's ability not only to localize explicit referents but also to reason over subtle visual cues and contextual relationships in complex scenes.

As shown in \cref{fig:sup-vis-wo-ref} (top-left), VGent successfully identifies the person wearing glasses in a densely crowded environment. 
Despite the glasses covering only a few pixels and the presence of numerous distractor individuals without glasses, the model accurately grounds the intended target. 
This demonstrates VGent’s strong sensitivity to fine-grained visual attributes and its capability to filter out semantically similar distractors.

Similarly, in \cref{fig:sup-vis-w-ref} (top-left), VGent effectively resolves a visual reference-conditioned multi-target query, detecting all people above the provided visual reference. 
The model succeeds even under occlusion and when some targets appear at a small scale due to being farther from the camera. 
These results illustrate VGent’s ability to integrate visual reference signals, reason over relational cues, and maintain stable grounding performance.

\begin{table}[th]
    \centering
    \caption{\textbf{Results on generalized referring expression segmentation (GRES) and reasoning segmentation (ReasonSeg).} We highlight the best performance in bold and underline the second best.}
    \resizebox{0.48\textwidth}{!}{
    \begin{tabular}{l c c c c c c}
        \toprule
        & \multicolumn{4}{c}{GRES} & \multicolumn{2}{c}{ReasonSeg} \\
        \cmidrule(lr){2-5}\cmidrule(lr){6-7}
        Model &  F1 & gIoU & cIoU & N-acc & gIoU & cIoU \\
        \midrule
        MagNet~\cite{chng2024mask}   & - & - & - & - & - & - \\
        Groundhog$_\text{7B}$~\cite{zhang2024groundhog}   & - & - & - & - & - & -\\
        GLaMM$_\text{7B, FT}$~\cite{rasheed2024glamm}   & - &  - & -  &- & -&- \\
        u-LLaVA$_\text{7B}$~\cite{xu2024ullava}   &- & -& -& -& -&- \\
        UNINEXT-H~\cite{yan2023universal}  & -& -& -& -& -&- \\
        PSALM$_\text{1.3B}$~\cite{zhang2024psalm}   & -&- &- & -& -&- \\
        LAVT~\cite{yang2022lavt}& - & 58.40 & 57.64 & 49.32 &- &- \\
        HDC~\cite{luo2024hdc}  & - & 68.28 & 65.42 & 63.38 & - & - \\
        ReLA~\cite{liu2023gres} & -  & 63.60 & 62.42 & 56.37  & 21.3 &- \\
        Seg-Zero~\cite{segzero} & - & - & - & - & 57.5  & 52.0 \\
        GSVA$_\text{13B, FT}$~\cite{xia2024gsva} & - & 70.04 & 66.38 & 66.02 & - & - \\
        SAM4MLLM$_\text{7B}$~\cite{chen2024sam4mllm}  & - & 71.86 & 67.83 & 66.08 & - &- \\
        LISA$_\text{13B, FT}$~\cite{lai2024lisa}& - & 65.24 & 63.96 & 57.49 & 61.3 &  62.2  \\
        RAS$_\text{13B}$~\cite{ores} & \underline{81.74} & \underline{74.64} & \bf{70.48} & \underline{69.05} & - &- \\
        \midrule
        VGent (Ours)  & \bf 82.91 & \bf 77.14 & \underline{69.33} & \bf 83.33 & \bf  62.2 & \bf 64.0\\
        \bottomrule
    \end{tabular}
    }
    \label{tab:sup-seg}
\end{table}

\section{Additional Quantitative Results}
\label{sec:additional_quant}

In \cref{tab:sup-seg}, we further report experimental results on 
generalized referring expression segmentation (GRES) evaluated on gRefCOCO val split
and Reasoning Segmentation (ReasonSeg) evaluated on the ReasonSeg test split. 
GRES~\cite{gref} involves an arbitrary number of targets, and ReasonSeg~\cite{lisa} evaluates grounding under complex and implicit language instructions. 
VGent achieves superior performance, demonstrating the robustness and generalization capability of our framework across diverse grounding scenarios. 
In particular, VGent achieves a substantial improvement in the GRES N-Acc metric—which evaluates whether the model hallucinates targets in non-target scenarios—surpassing the previous state-of-the-art RAS$_\text{13B}$~\cite{ores} by \textbf{+14.28\%}. 
This result highlights the faithfulness of VGent and its significantly reduced tendency to hallucinate outputs.

\section{Ablation on Upper Bounds}
\begin{table}[th]
    \centering
    \vspace{-2mm}
    \caption{{Oracle selection for upper-bound performance on Omnimodal Referring Expression Segmentation (ORES).} 
}
\resizebox{0.48\textwidth}{!}{
    \begin{tabular}{l c c c}
        \toprule
        \multirow{3}{*}{{Model}} & \multicolumn{3}{c}{Overall} \\
        \cmidrule(lr){2-4}
         & F1 & gIoU & cIoU \\
        \midrule
        VGent (Ours)  & 71.47 & 68.42 &  75.28 \\
        UPN~\cite{upn} (Oracle) & 91.27 & 79.97 & 81.40 \\
        UPN~\cite{upn} + GLEE~\cite{glee} (Oracle)  & 94.68 & 84.05 & 85.00 \\
        UPN~\cite{upn} + GLEE~\cite{glee} + SAM~\cite{sam2} (Oracle) & \bf 95.38 & \bf 86.20 & \bf 88.45 \\
        \bottomrule
    \end{tabular}}
    \vspace{-2mm}
    \label{tab:ores-oracle}
\end{table}

We evaluate how different detector combinations affect the upper-bound performance of VGent by applying oracle selection on ORES. For F1, we run Hungarian Matching between the grouth truth boxes and proposed boxes, and retain proposals whose IoU exceeds 0.5; for gIoU and cIoU, we keep proposals whose IoA exceeds 0.6. As shown in Table \ref{tab:ores-oracle}, different detectors provide complementary proposals that jointly increase coverage of the ground-truth boxes, thereby raising the achievable upper bound of VGent’s performance.

\section{Details of Implementation}
\label{sec:impl_details}

\paragraph{QuadThinker}
For the QuadThinker component used to initialize VGent's encoder, 
we perform GRPO training for one epoch based on Qwen2.5-VL-7B~\cite{qwen2.5vl} 
using MaskGroups-HQ~\cite{ores} and VisionReasoner-7K~\cite{visionreasoner}, 
with a batch size of 16 and a learning rate of 1e-6.

\paragraph{Learnable Query}
Inspired by SegVG~\cite{segvg}, we use multiple learnable queries to benefit proposal selection through self-attention within each decoder layer which propagates the global target information. 
Empirically, we find that using 10 learnable queries yields the best performance, where 5 queries are used to regress the number of targets and 5 are used to regress the number of positive proposals.

\paragraph{Visual Reference}
MaskGroups-HQ~\cite{ores} provides visual references in the form of segmentation masks. To integrate these visual references into the language query, we convert each mask into a bounding box. Specifically, we compute the minimum and maximum (x,y) coordinates that tightly enclose the mask, resize the resulting box to the resolution of the model’s image input, and round all coordinates to integers. We then replace the placeholder token \texttt{\textless mask-ref\textgreater} in the textual query with this coordinate list. For example, the query \textit{“the woman wearing a skirt behind the left side of \texttt{\textless mask-ref\textgreater}”} becomes \textit{“the woman wearing a skirt behind the left side of [50, 490, 120, 637]”.}

\paragraph{Training on ORES}
For experiments on ORES, which follows the evaluation split of MaskGroups-HQ~\cite{ores}, 
we combine proposals from UPN~\cite{upn}, SAM~\cite{sam2}, and GLEE~\cite{glee} during training. 
We first train on Objects365~\cite{objects365} for 16K steps using 6 nodes (each with 8$\times$A100-80G GPUs), 
with a per-GPU batch size of 1 and gradient accumulation of 2. 
We then train on the mixed dataset of Objects365~\cite{objects365} and MaskGroups-2M~\cite{ores}, 
sampled with the 0.3 and 0.7 ratio of them under the same configuration. 
Finally, we train on the MaskGroups-HQ~\cite{ores} training split for 48K steps using 1 node of 8$\times$A100-80G GPUs.
The BCE loss is weighted by 1 and the L1 loss by 10. 
We use a learning rate of 2e-5 with linear decay.
For box-aware label, proposals with IoU $>$ 0.6 are treated as positives and all others as negatives. 
For mask-aware label, we assign positives using IoA $>$ 0.6. 
All images are resized to $840 \times 840$ resolution.

\paragraph{Training on REC}
For REC experiments, we follow RAS~\cite{ores} to further fine-tune on all training splits of RefCOCO, RefCOCO+, and RefCOCOg for 48K steps using 1 node of 8$\times$A100-80G GPUs. 

\paragraph{Training on GRES and ReasonSeg}
For experiments on GRES and ReasonSeg, we fine-tune the checkpoint obtained after pre-training on Objects365~\cite{objects365} and MaskGroups-2M~\cite{ores}. 
During fine-tuning, we reweight the loss for mask-aware labels by a factor of $1 + \text{IoA}$ for each proposal on GRES.
All fine-tuning experiments are conducted on their respective training splits for 48K steps using a single node with 8$\times$A100-80G GPUs. 
We report results based on the best-performing checkpoint and outputs.

\paragraph{Inference}
We use UPN~\cite{upn}, SAM~\cite{sam2}, and GLEE~\cite{glee} for both training and inference, and for all inference-time speed measurements. 
The runtime consists of 0.696 seconds for VGent's encoder--decoder, 0.263 seconds for UPN, 
0.213 seconds for GLEE, and 1.154 seconds for SAM.

\paragraph{Ablation Studies}
For ablation experiments, QuadThinker is further trained for four additional epochs when being integrated into VGent. 
While this extended training does not improve QuadThinker's performance, 
it consistently yields better overall performance for VGent. 
All ablation studies are conducted on a single node with 8$\times$A100-80G GPUs.

\paragraph{Qwen3-VL Evaluation}
Following the official GitHub instructions of Qwen3-VL~\cite{qwen3}, we use the prompt:
\textit{``Locate \{Question\}, output the bbox coordinates using JSON format.''}, where \textit{\{Question\}} is replaced by the language query input.
For consistency with our implementation, the input image is resized to a resolution of 840 $\times$ 840. 
Qwen3-VL outputs bounding boxes in a normalized format, where each coordinate is represented as a relative value multiplied by 1000. 
During post-processing, we divide the predicted values by 1000 and scale them by the image resolution to recover the absolute bounding box coordinates.

%% file: main.bib
@String(CVPR= {IEEE Conf. Comput. Vis. Pattern Recog.})

@String(ECCV= {Eur. Conf. Comput. Vis.})

@String(ICLR = {Int. Conf. Learn. Represent.})

@String(CVPR  = {CVPR})

@String(ECCV  = {ECCV})

@String(ICLR  = {ICLR})

@article{qi2022high,
  title={High-quality entity segmentation},
  author={Qi, Lu and Kuen, Jason and Guo, Weidong and Shen, Tiancheng and Gu, Jiuxiang and Jia, Jiaya and Lin, Zhe and Yang, Ming-Hsuan},
  journal={arXiv preprint arXiv:2211.05776},
  year={2022}
}

@article{rahimi2025explanations,
  title={Explanations of Large Language Models Explain Language Representations in the Brain},
  author={Rahimi, Maryam and Yaghoobzadeh, Yadollah and Daliri, Mohammad Reza},
  journal={arXiv preprint arXiv:2502.14671},
  year={2025}
}

@article{oota2023joint,
  title={Joint processing of linguistic properties in brains and language models},
  author={Oota, SubbaReddy and Gupta, Manish and Toneva, Mariya},
  journal={Advances in Neural Information Processing Systems},
  volume={36},
  pages={18001--18014},
  year={2023}
}

@article{fan2024not,
  title={Not all layers of llms are necessary during inference},
  author={Fan, Siqi and Jiang, Xin and Li, Xiang and Meng, Xuying and Han, Peng and Shang, Shuo and Sun, Aixin and Wang, Yequan and Wang, Zhongyuan},
  journal={arXiv preprint arXiv:2403.02181},
  year={2024}
}

@article{vfl,
  title={Vision Function Layer in Multimodal LLMs},
  author={Shi, Cheng and Yu, Yizhou and Yang, Sibei},
  journal={arXiv preprint arXiv:2509.24791},
  year={2025}
}

@inproceedings{gan2017vqs,
  title={Vqs: Linking segmentations to questions and answers for supervised attention in vqa and question-focused semantic segmentation},
  author={Gan, Chuang and Li, Yandong and Li, Haoxiang and Sun, Chen and Gong, Boqing},
  booktitle={Proceedings of the IEEE international conference on computer vision},
  pages={1811--1820},
  year={2017}
}

@inproceedings{sam3,
  title={SAM 3: Segment Anything with Concepts},
  booktitle={ICLR 2026 Conference Submission (under review)},
  year={2025},
  url={https://openreview.net/forum?id=r35clVtGzw}
}

@inproceedings{objects365,
  title={Objects365: A large-scale, high-quality dataset for object detection},
  author={Shao, Shuai and Li, Zeming and Zhang, Tianyuan and Peng, Chao and Yu, Gang and Zhang, Xiangyu and Li, Jing and Sun, Jian},
  booktitle={Proceedings of the IEEE/CVF international conference on computer vision},
  pages={8430--8439},
  year={2019}
}

@inproceedings{glee,
  title={General object foundation model for images and videos at scale},
  author={Wu, Junfeng and Jiang, Yi and Liu, Qihao and Yuan, Zehuan and Bai, Xiang and Bai, Song},
  booktitle={Proceedings of the IEEE/CVF Conference on Computer Vision and Pattern Recognition},
  pages={3783--3795},
  year={2024}
}

@article{sam2,
  title={Sam 2: Segment anything in images and videos},
  author={Ravi, Nikhila and Gabeur, Valentin and Hu, Yuan-Ting and Hu, Ronghang and Ryali, Chaitanya and Ma, Tengyu and Khedr, Haitham and R{\"a}dle, Roman and Rolland, Chloe and Gustafson, Laura and others},
  journal={arXiv preprint arXiv:2408.00714},
  year={2024}
}

@article{upn,
  title={Chatrex: Taming multimodal llm for joint perception and understanding},
  author={Jiang, Qing and Luo, Gen and Yang, Yuqin and Xiong, Yuda and Chen, Yihao and Zeng, Zhaoyang and Ren, Tianhe and Zhang, Lei},
  journal={arXiv preprint arXiv:2411.18363},
  year={2024}
}

@article{internvl3.5,
  title={Internvl3. 5: Advancing open-source multimodal models in versatility, reasoning, and efficiency},
  author={Wang, Weiyun and Gao, Zhangwei and Gu, Lixin and Pu, Hengjun and Cui, Long and Wei, Xingguang and Liu, Zhaoyang and Jing, Linglin and Ye, Shenglong and Shao, Jie and others},
  journal={arXiv preprint arXiv:2508.18265},
  year={2025}
}

@article{internvl3,
  title={Internvl3: Exploring advanced training and test-time recipes for open-source multimodal models},
  author={Zhu, Jinguo and Wang, Weiyun and Chen, Zhe and Liu, Zhaoyang and Ye, Shenglong and Gu, Lixin and Tian, Hao and Duan, Yuchen and Su, Weijie and Shao, Jie and others},
  journal={arXiv preprint arXiv:2504.10479},
  year={2025}
}

@article{gemini,
  title={Gemini 2.5: Pushing the frontier with advanced reasoning, multimodality, long context, and next generation agentic capabilities},
  author={Comanici, Gheorghe and Bieber, Eric and Schaekermann, Mike and Pasupat, Ice and Sachdeva, Noveen and Dhillon, Inderjit and Blistein, Marcel and Ram, Ori and Zhang, Dan and Rosen, Evan and others},
  journal={arXiv preprint arXiv:2507.06261},
  year={2025}
}

@article{one-peace,
author    = {Peng Wang and Shijie Wang and Junyang Lin and Shuai Bai and Xiaohuan Zhou and Jingren Zhou and Xinggang Wang and Chang Zhou},
title     = {One-peace: Exploring one general representation model toward unlimited modalities},
journal   = {arXiv:2305.11172},
year      = {2023}
}

@inproceedings{grounding_dino,
author    = {Shilong Liu and Zhaoyang Zeng and Tianhe Ren and Feng Li and Hao Zhang and Jie Yang and Qing Jiang and Chunyuan Li and Jianwei Yang and Hang Su and others},
title     = {Grounding dino: Marrying dino with grounded pre-training for open-set object detection},
booktitle = {European Conference on Computer Vision},
pages     = {38--55},
publisher = {Springer},
year      = {2025}
}

@misc{qwen3,
      title={Qwen3 Technical Report}, 
      author={Qwen Team},
      year={2025},
      eprint={2505.09388},
      archivePrefix={arXiv},
      primaryClass={cs.CL},
      url={https://arxiv.org/abs/2505.09388}, 
}

@inproceedings{liu2023gres,
  title={{GRES}: Generalized referring expression segmentation},
  author={Liu, Chang and Ding, Henghui and Jiang, Xudong},
  booktitle={CVPR},
  year={2023}
}

@inproceedings{yang2022lavt,
  title={{LAVT}: Language-aware vision transformer for referring image segmentation},
  author={Yang, Zhao and Wang, Jiaqi and Tang, Yansong and Chen, Kai and Zhao, Hengshuang and Torr, Philip H.S.},
  booktitle={CVPR},
  year={2022}
}

@inproceedings{yan2023universal,
  title={Universal instance perception as object discovery and retrieval},
  author={Yan, Bin and Jiang, Yi and Wu, Jiannan and Wang, Dong and Luo, Ping and Yuan, Zehuan and Lu, Huchuan},
  booktitle={CVPR},
  year={2023}
}

@inproceedings{lai2024lisa,
  title={{LISA}: Reasoning segmentation via large language model},
  author={Lai, Xin and Tian, Zhuotao and Chen, Yukang and Li, Yanwei and Yuan, Yuhui and Liu, Shu and Jia, Jiaya},
  booktitle={CVPR},
  year={2024}
}

@inproceedings{rasheed2024glamm,
  title={{GLaMM}: Pixel grounding large multimodal model},
  author={Rasheed, Hanoona and Maaz, Muhammad and Shaji, Sahal and Shaker, Abdelrahman and Khan, Salman and Cholakkal, Hisham and Anwer, Rao M. and Xing, Eric and Yang, Ming-Hsuan and Khan, Fahad S.},
  booktitle={CVPR},
  year={2024}
}

@inproceedings{zhang2024groundhog,
  title={{Groundhog}: Grounding large language models to holistic segmentation},
  author={Zhang, Yichi and Ma, Ziqiao and Gao, Xiaofeng and Shakiah, Suhaila and Gao, Qiaozi and Chai, Joyce},
  booktitle={CVPR},
  year={2024}
}

@inproceedings{chng2024mask,
  title={Mask grounding for referring image segmentation},
  author={Chng, Yong Xien and Zheng, Henry and Han, Yizeng and Qiu, Xuchong and Huang, Gao},
  booktitle={CVPR},
  year={2024}
}

@inproceedings{xia2024gsva,
  title={{GSVA}: Generalized segmentation via multimodal large language models},
  author={Xia, Zhuofan and Han, Dongchen and Han, Yizeng and Pan, Xuran and Song, Shiji and Huang, Gao},
  booktitle={CVPR},
  year={2024}
}

@inproceedings{xu2024ullava,
  title={{u-LLaVA}: Unifying Multi-Modal Tasks via Large Language Model},
  author={Xu, Jinjin and Xu, Liwu and Yang, Yuzhe and Li, Xiang and Wang, Fanyi and Xie, Yanchun and Huang, Yi-Jie and Li, Yaqian},
  booktitle={ECAI},
  year={2024}
}

@inproceedings{chen2024sam4mllm,
  title={{SAM4MLLM}: Enhance Multi-Modal Large Language Model for Referring Expression Segmentation},
  author={Chen, Yi-Chia and Li, Wei-Hua and Sun, Cheng and Wang, Yu-Chiang Frank and Chen, Chu-Song},
  booktitle={ECCV},
  year={2024}
}

@inproceedings{zhang2024psalm,
  title={{PSALM}: Pixelwise segmentation with large multi-modal model},
  author={Zhang, Zheng and Ma, Yeyao and Zhang, Enming and Bai, Xiang},
  booktitle={ECCV},
  year={2024}
}

@article{luo2024hdc,
  title={{HDC}: Hierarchical Semantic Decoding with Counting Assistance for Generalized Referring Expression Segmentation},
  author={Luo, Zhuoyan and Wu, Yinghao and Liu, Yong and Xiao, Yicheng and Zhang, Xiao-Ping and Yang, Yujiu},
  journal={arXiv preprint arXiv:2405.15658},
  year={2024}
}

@misc{grpo,
      title={DeepSeekMath: Pushing the Limits of Mathematical Reasoning in Open Language Models}, 
      author={Zhihong Shao and Peiyi Wang and Qihao Zhu and Runxin Xu and Junxiao Song and Xiao Bi and Haowei Zhang and Mingchuan Zhang and Y. K. Li and Y. Wu and Daya Guo},
      year={2024},
      eprint={2402.03300},
      archivePrefix={arXiv},
      primaryClass={cs.CL},
      url={https://arxiv.org/abs/2402.03300}, 
}

@article{text4seg,
  title={Text4seg: Reimagining image segmentation as text generation},
  author={Lan, Mengcheng and Chen, Chaofeng and Zhou, Yue and Xu, Jiaxing and Ke, Yiping and Wang, Xinjiang and Feng, Litong and Zhang, Wayne},
  journal={arXiv preprint arXiv:2410.09855},
  year={2024}
}

@article{visionreasoner,
  title={VisionReasoner: Unified Visual Perception and Reasoning via Reinforcement Learning},
  author={Liu, Yuqi and Qu, Tianyuan and Zhong, Zhisheng and Peng, Bohao and Liu, Shu and Yu, Bei and Jia, Jiaya},
  journal={arXiv preprint arXiv:2505.12081},
  year={2025}
}

@article{segzero,
  title={Seg-zero: Reasoning-chain guided segmentation via cognitive reinforcement},
  author={Liu, Yuqi and Peng, Bohao and Zhong, Zhisheng and Yue, Zihao and Lu, Fanbin and Yu, Bei and Jia, Jiaya},
  journal={arXiv preprint arXiv:2503.06520},
  year={2025}
}

@misc{vicuna,
    title = {Vicuna: An Open-Source Chatbot Impressing GPT-4 with 90\%* ChatGPT Quality},
    url = {https://lmsys.org/blog/2023-03-30-vicuna/},
    author = {Chiang, Wei-Lin and Li, Zhuohan and Lin, Zi and Sheng, Ying and Wu, Zhanghao and Zhang, Hao and Zheng, Lianmin and Zhuang, Siyuan and Zhuang, Yonghao and Gonzalez, Joseph E. and Stoica, Ion and Xing, Eric P.},
    month = {March},
    year = {2023}
}

@article{llama2,
  title={Llama 2: Open foundation and fine-tuned chat models},
  author={Touvron, Hugo and Martin, Louis and Stone, Kevin and Albert, Peter and Almahairi, Amjad and Babaei, Yasmine and Bashlykov, Nikolay and Batra, Soumya and Bhargava, Prajjwal and Bhosale, Shruti and others},
  journal={arXiv preprint arXiv:2307.09288},
  year={2023}
}

@article{qwen2,
  title={Qwen2-vl: Enhancing vision-language model's perception of the world at any resolution},
  author={Wang, Peng and Bai, Shuai and Tan, Sinan and Wang, Shijie and Fan, Zhihao and Bai, Jinze and Chen, Keqin and Liu, Xuejing and Wang, Jialin and Ge, Wenbin and others},
  journal={arXiv preprint arXiv:2409.12191},
  year={2024}
}

@article{hallucination,
  title={Hallucination of multimodal large language models: A survey},
  author={Bai, Zechen and Wang, Pichao and Xiao, Tianjun and He, Tong and Han, Zongbo and Zhang, Zheng and Shou, Mike Zheng},
  journal={arXiv preprint arXiv:2404.18930},
  year={2024}
}

@inproceedings{visualhallucination,
    title = "Visual Hallucinations of Multi-modal Large Language Models",
    author = "Huang, Wen  and
      Liu, Hongbin  and
      Guo, Minxin  and
      Gong, Neil",
    editor = "Ku, Lun-Wei  and
      Martins, Andre  and
      Srikumar, Vivek",
    booktitle = "Findings of the Association for Computational Linguistics: ACL 2024",
    month = aug,
    year = "2024",
    address = "Bangkok, Thailand",
    publisher = "Association for Computational Linguistics",
    url = "https://aclanthology.org/2024.findings-acl.573/",
    doi = "10.18653/v1/2024.findings-acl.573",
    pages = "9614--9631"
}

@article{pope,
  title={Evaluating object hallucination in large vision-language models},
  author={Li, Yifan and Du, Yifan and Zhou, Kun and Wang, Jinpeng and Zhao, Wayne Xin and Wen, Ji-Rong},
  journal={arXiv preprint arXiv:2305.10355},
  year={2023}
}

@article{ores,
  title={Refer to Anything with Vision-Language Prompts},
  author={Cao, Shengcao and Wei, Zijun and Kuen, Jason and Liu, Kangning and Zhang, Lingzhi and Gu, Jiuxiang and Jung, HyunJoon and Gui, Liang-Yan and Wang, Yu-Xiong},
  journal={arXiv preprint arXiv:2506.05342},
  year={2025}
}

@inproceedings{gref,
  title={Gres: Generalized referring expression segmentation},
  author={Liu, Chang and Ding, Henghui and Jiang, Xudong},
  booktitle={Proceedings of the IEEE/CVF conference on computer vision and pattern recognition},
  pages={23592--23601},
  year={2023}
}

@inproceedings{refcocog,
  title={Generation and comprehension of unambiguous object descriptions},
  author={Mao, Junhua and Huang, Jonathan and Toshev, Alexander and Camburu, Oana and Yuille, Alan L and Murphy, Kevin},
  booktitle={Proceedings of the IEEE conference on computer vision and pattern recognition},
  pages={11--20},
  year={2016}
}

@inproceedings{refcoco,
  title={Modeling context in referring expressions},
  author={Yu, Licheng and Poirson, Patrick and Yang, Shan and Berg, Alexander C and Berg, Tamara L},
  booktitle={European conference on computer vision},
  pages={69--85},
  year={2016},
  organization={Springer}
}

@article{lumen,
  title={Lumen: Unleashing versatile vision-centric capabilities of large multimodal models},
  author={Jiao, Yang and Chen, Shaoxiang and Jie, Zequn and Chen, Jingjing and Ma, Lin and Jiang, Yu-Gang},
  journal={Advances in Neural Information Processing Systems},
  volume={37},
  pages={81461--81488},
  year={2024}
}

@article{omgllava,
  title={Omg-llava: Bridging image-level, object-level, pixel-level reasoning and understanding},
  author={Zhang, Tao and Li, Xiangtai and Fei, Hao and Yuan, Haobo and Wu, Shengqiong and Ji, Shunping and Loy, Chen Change and Yan, Shuicheng},
  journal={Advances in neural information processing systems},
  volume={37},
  pages={71737--71767},
  year={2024}
}

@inproceedings{glamm,
  title={Glamm: Pixel grounding large multimodal model},
  author={Rasheed, Hanoona and Maaz, Muhammad and Shaji, Sahal and Shaker, Abdelrahman and Khan, Salman and Cholakkal, Hisham and Anwer, Rao M and Xing, Eric and Yang, Ming-Hsuan and Khan, Fahad S},
  booktitle={Proceedings of the IEEE/CVF Conference on Computer Vision and Pattern Recognition},
  pages={13009--13018},
  year={2024}
}

@article{visionllm2,
  title={Visionllm v2: An end-to-end generalist multimodal large language model for hundreds of vision-language tasks},
  author={Wu, Jiannan and Zhong, Muyan and Xing, Sen and Lai, Zeqiang and Liu, Zhaoyang and Chen, Zhe and Wang, Wenhai and Zhu, Xizhou and Lu, Lewei and Lu, Tong and others},
  journal={Advances in Neural Information Processing Systems},
  volume={37},
  pages={69925--69975},
  year={2024}
}

@inproceedings{pixellm,
  title={Pixellm: Pixel reasoning with large multimodal model},
  author={Ren, Zhongwei and Huang, Zhicheng and Wei, Yunchao and Zhao, Yao and Fu, Dongmei and Feng, Jiashi and Jin, Xiaojie},
  booktitle={Proceedings of the IEEE/CVF Conference on Computer Vision and Pattern Recognition},
  pages={26374--26383},
  year={2024}
}

@inproceedings{lisa,
  title={Lisa: Reasoning segmentation via large language model},
  author={Lai, Xin and Tian, Zhuotao and Chen, Yukang and Li, Yanwei and Yuan, Yuhui and Liu, Shu and Jia, Jiaya},
  booktitle={Proceedings of the IEEE/CVF Conference on Computer Vision and Pattern Recognition},
  pages={9579--9589},
  year={2024}
}

@article{chatscene,
  title={Chat-scene: Bridging 3d scene and large language models with object identifiers},
  author={Huang, Haifeng and Chen, Yilun and Wang, Zehan and Huang, Rongjie and Xu, Runsen and Wang, Tai and Liu, Luping and Cheng, Xize and Zhao, Yang and Pang, Jiangmiao and others},
  journal={Advances in Neural Information Processing Systems},
  volume={37},
  pages={113991--114017},
  year={2024}
}

@inproceedings{groma,
  title={Groma: Localized visual tokenization for grounding multimodal large language models},
  author={Ma, Chuofan and Jiang, Yi and Wu, Jiannan and Yuan, Zehuan and Qi, Xiaojuan},
  booktitle={European Conference on Computer Vision},
  pages={417--435},
  year={2024},
  organization={Springer}
}

@article{lmmdet,
  title={LMM-Det: Make Large Multimodal Models Excel in Object Detection},
  author={Li, Jincheng and Xie, Chunyu and Ao, Ji and Leng, Dawei and Yin, Yuhui},
  journal={arXiv preprint arXiv:2507.18300},
  year={2025}
}

@article{ferret,
  title={Ferret: Refer and ground anything anywhere at any granularity},
  author={You, Haoxuan and Zhang, Haotian and Gan, Zhe and Du, Xianzhi and Zhang, Bowen and Wang, Zirui and Cao, Liangliang and Chang, Shih-Fu and Yang, Yinfei},
  journal={arXiv preprint arXiv:2310.07704},
  year={2023}
}

@article{ferret2,
  title={Ferret-v2: An improved baseline for referring and grounding with large language models},
  author={Zhang, Haotian and You, Haoxuan and Dufter, Philipp and Zhang, Bowen and Chen, Chen and Chen, Hong-You and Fu, Tsu-Jui and Wang, William Yang and Chang, Shih-Fu and Gan, Zhe and others},
  journal={arXiv preprint arXiv:2404.07973},
  year={2024}
}

@article{kosmos2,
  title={Kosmos-2: Grounding multimodal large language models to the world},
  author={Peng, Zhiliang and Wang, Wenhui and Dong, Li and Hao, Yaru and Huang, Shaohan and Ma, Shuming and Wei, Furu},
  journal={arXiv preprint arXiv:2306.14824},
  year={2023}
}

@article{shikra,
  title={Shikra: Unleashing multimodal llm's referential dialogue magic},
  author={Chen, Keqin and Zhang, Zhao and Zeng, Weili and Zhang, Richong and Zhu, Feng and Zhao, Rui},
  journal={arXiv preprint arXiv:2306.15195},
  year={2023}
}

@inproceedings{vltvg,
  title={Improving visual grounding with visual-linguistic verification and iterative reasoning},
  author={Yang, Li and Xu, Yan and Yuan, Chunfeng and Liu, Wei and Li, Bing and Hu, Weiming},
  booktitle={Proceedings of the IEEE/CVF Conference on Computer Vision and Pattern Recognition},
  pages={9499--9508},
  year={2022}
}

@inproceedings{transvg,
  title={Transvg: End-to-end visual grounding with transformers},
  author={Deng, Jiajun and Yang, Zhengyuan and Chen, Tianlang and Zhou, Wengang and Li, Houqiang},
  booktitle={Proceedings of the IEEE/CVF international conference on computer vision},
  pages={1769--1779},
  year={2021}
}

@article{qwen2.5vl,
  title={Qwen2. 5-vl technical report},
  author={Bai, Shuai and Chen, Keqin and Liu, Xuejing and Wang, Jialin and Ge, Wenbin and Song, Sibo and Dang, Kai and Wang, Peng and Wang, Shijie and Tang, Jun and others},
  journal={arXiv preprint arXiv:2502.13923},
  year={2025}
}

@inproceedings{llava1.5,
  title={Improved baselines with visual instruction tuning},
  author={Liu, Haotian and Li, Chunyuan and Li, Yuheng and Lee, Yong Jae},
  booktitle={Proceedings of the IEEE/CVF conference on computer vision and pattern recognition},
  pages={26296--26306},
  year={2024}
}

@misc{segvg,
      title={SegVG: Transferring Object Bounding Box to Segmentation for Visual Grounding}, 
      author={Weitai Kang and Gaowen Liu and Mubarak Shah and Yan Yan},
      year={2024},
      eprint={2407.03200},
      archivePrefix={arXiv},
      primaryClass={cs.CV},
      url={https://arxiv.org/abs/2407.03200}, 
}

@misc{intent3d,
      title={Intent3D: 3D Object Detection in RGB-D Scans Based on Human Intention}, 
      author={Weitai Kang and Mengxue Qu and Jyoti Kini and Yunchao Wei and Mubarak Shah and Yan Yan},
      year={2025},
      eprint={2405.18295},
      archivePrefix={arXiv},
      primaryClass={cs.CV},
      url={https://arxiv.org/abs/2405.18295}, 
}

@misc{actress,
      title={ACTRESS: Active Retraining for Semi-supervised Visual Grounding}, 
      author={Weitai Kang and Mengxue Qu and Yunchao Wei and Yan Yan},
      year={2024},
      eprint={2407.03251},
      archivePrefix={arXiv},
      primaryClass={cs.CV},
      url={https://arxiv.org/abs/2407.03251}, 
}

@misc{robin3d,
      title={Robin3D: Improving 3D Large Language Model via Robust Instruction Tuning}, 
      author={Weitai Kang and Haifeng Huang and Yuzhang Shang and Mubarak Shah and Yan Yan},
      year={2025},
      eprint={2410.00255},
      archivePrefix={arXiv},
      primaryClass={cs.AI},
      url={https://arxiv.org/abs/2410.00255}, 
}

@misc{attbalance,
      title={Visual Grounding with Attention-Driven Constraint Balancing}, 
      author={Weitai Kang and Luowei Zhou and Junyi Wu and Changchang Sun and Yan Yan},
      year={2024},
      eprint={2407.03243},
      archivePrefix={arXiv},
      primaryClass={cs.CV},
      url={https://arxiv.org/abs/2407.03243}, 
}

@misc{guirlvg,
      title={GuirlVG: Incentivize GUI Visual Grounding via Empirical Exploration on Reinforcement Learning}, 
      author={Weitai Kang and Bin Lei and Gaowen Liu and Caiwen Ding and Yan Yan},
      year={2025},
      eprint={2508.04389},
      archivePrefix={arXiv},
      primaryClass={cs.AI},
      url={https://arxiv.org/abs/2508.04389}, 
}

@misc{infantagent-next,
      title={InfantAgent-Next: A Multimodal Generalist Agent for Automated Computer Interaction}, 
      author={Bin Lei and Weitai Kang and Zijian Zhang and Winson Chen and Xi Xie and Shan Zuo and Mimi Xie and Ali Payani and Mingyi Hong and Yan Yan and Caiwen Ding},
      year={2025},
      eprint={2505.10887},
      archivePrefix={arXiv},
      primaryClass={cs.AI},
      url={https://arxiv.org/abs/2505.10887}, 
}

@misc{expvg,
      title={ExpVG: Investigating the Design Space of Visual Grounding in Multimodal Large Language Model}, 
      author={Weitai Kang and Weiming Zhuang and Zhizhong Li and Yan Yan and Lingjuan Lyu},
      year={2025},
      eprint={2508.08066},
      archivePrefix={arXiv},
      primaryClass={cs.CV},
      url={https://arxiv.org/abs/2508.08066}, 
}
